\documentclass{IEEEtran}
\usepackage{graphicx}
\usepackage{amsmath}
\usepackage{cite}
\usepackage{booktabs}
\usepackage{multirow}
\usepackage{caption}
\usepackage{subcaption}
\usepackage{xcolor,colortbl}
\usepackage{svg}
\usepackage{amsfonts}
\usepackage{amssymb}

\title{ECOR: Explainable CLIP for Object Recognition}
\author{Ali Rasekh$^{*1}$, Sepehr Kazemi Ranjbar$^{*2}$, Milad Heidari$^{2}$, Wolfgang Nejdl$^{1}$\\
    $^{1}$\{ali.rasekh, nejdl\}@l3s.de\\
    $^{2}$\{sepehrkazemi9, milad.h9999\}@gmail.com\\
    }

\newtheorem{definition}{Definition}
\newcommand{\cb}[1]{\left\{#1\right\}}
\newcommand{\pran}[1]{\left(#1\right)}

\definecolor{Gray}{gray}{0.85}
\newcolumntype{a}{>{\columncolor{Gray}}l}

\begin{document}
\maketitle

\def\thefootnote{*}\footnotetext{These authors contributed equally to this work.}

\begin{abstract}
Large Vision Language Models (VLMs), such as CLIP, have significantly contributed to various computer vision tasks, including object recognition and object detection. Their open vocabulary feature enhances their value. However, their black-box nature and lack of explainability in predictions make them less trustworthy in critical domains. Recently, some work has been done to force VLMs to provide reasonable rationales for object recognition, but this often comes at the expense of classification accuracy. In this paper, we first propose a mathematical definition of explainability in the object recognition task based on the joint probability distribution of categories and rationales, then leverage this definition to fine-tune CLIP in an explainable manner. Through evaluations of different datasets, our method demonstrates state-of-the-art performance in explainable classification. Notably, it excels in zero-shot settings, showcasing its adaptability. This advancement improves explainable object recognition, enhancing trust across diverse applications. The code will be made available online upon publication.
\end{abstract}

\section{Introduction}

\label{sec:intro}

Large vision language models (VLMs), like CLIP \cite{pmlr-v139-radford21a}, have revolutionized image classification. Despite the advancements made by earlier deep classification models such as AlexNet \cite{krizhevsky2012imagenet} and ResNet \cite{He_2016_CVPR}, their capacity to handle open-vocabulary classification contributes significantly to their adaptability across various domains. Furthermore, by fine-tuning them on specific datasets, they achieve remarkable accuracy \cite{gao2024clip, zhou2022learning, jia2022visual}. However, a fundamental challenge persists—their "black box" nature makes it challenging to comprehend why they classify images into specific categories. This lack of explainability presents significant obstacles in domains that require trust and accountability, such as healthcare \cite{jin2022explainable, singh2020explainable}, autonomous vehicles \cite{dong2022development, zablocki2022explainability}, and legal systems \cite{branting2021scalable, de2022explainable}.

\begin{figure}[tb]
    \centering
    \includegraphics[width=\linewidth]{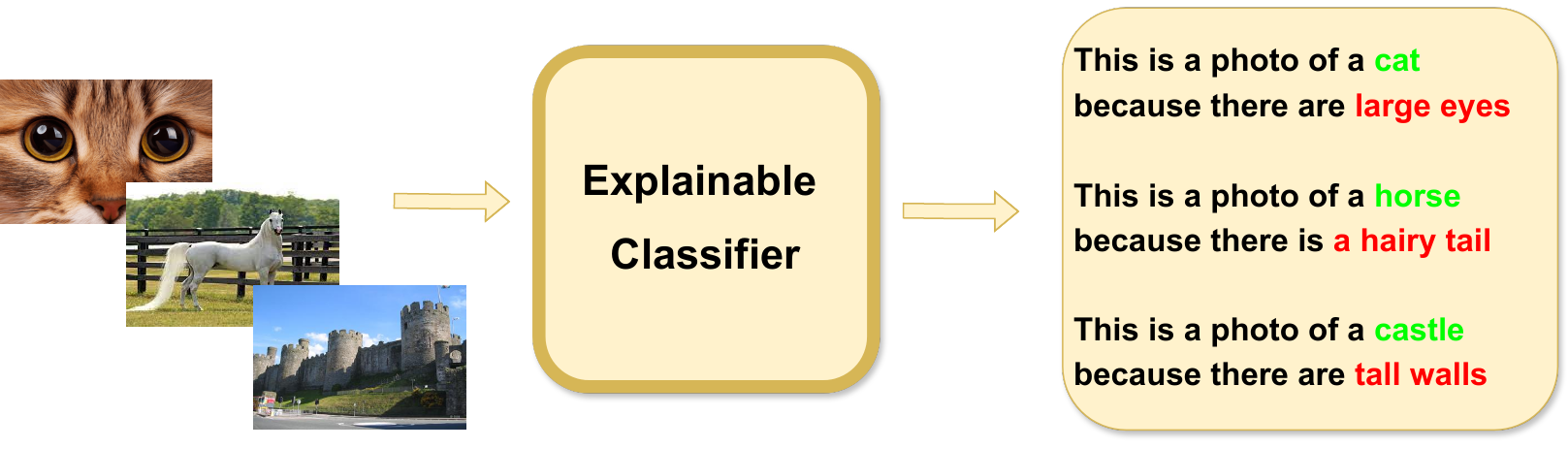}
      \caption{Our definition of explainability involves the effective utilization of true and relevant visual attributes, represented as text rationales, for object recognition and class prediction.}
    \label{fig:1}
\end{figure}

To address this issue, we require models that surpass mere prediction accuracy and offer meaningful explanations for their classifications. These meaningful explanations are known as rationales. In an image, there can be multiple rationales that lead us to identify a category. Most of the time, rationales are simpler and easier to understand, allowing both humans and deep neural networks to recognize them more accurately than categories alone. Thus, the problem lies in compelling VLMs to provide useful rationales for their class predictions. Figure \ref{fig:1} illustrates this process. Three images are inputted into an Explainable Classifier, categorizing them with corresponding rationales. For instance, tall walls serve as a significant clue for identifying a castle.

Traditional methods such as saliency maps \cite{selvaraju2017grad} focus on identifying influential image regions but often struggle to capture the broader reasoning process of complex DNNs. Recent advancements in Vision Language Models (VLMs) like CLIP offer promising steps toward explainability in these models. CLIP \cite{pmlr-v139-radford21a} is a contrastive vision language pre-training model trained on 400 million (image-caption) pairs sourced from across the internet. Despite its impressive classification accuracy and zero-shot performance, CLIP encounters challenges in providing useful rationales for its predictions, as demonstrated in Section \ref{sec:4} and also by \cite{menon2022visual, mao2023doubly}. In efforts to enhance CLIP's explainability, Menon \cite{menon2022visual} leveraged GPT3 \cite{NEURIPS2020_1457c0d6} to generate descriptive features (rationales) for each category, which were then fed along with categories into CLIP. Their approach yielded improved accuracy on various datasets compared to vanilla CLIP, indicating that providing rationales can enhance CLIP's performance. However, the rationales generated by GPT3 may not be present in all images of a category, and their evaluation solely based on category accuracy does not ensure the utilization of the correct rationales for category prediction. To address these limitations, Mao \cite{mao2023doubly} introduced valuable datasets comprising (image, category, rationale) triplets and proposed a new benchmark for explainable class prediction, requiring CLIP to predict both the category and rationale for an image. However, their approach significantly dropped classification accuracy when CLIP had to predict categories and provide useful rationales simultaneously.

Previous works have explored various approaches to achieve explainability, yet the primary challenge of maintaining prediction accuracy while providing useful rationales remains unresolved. In this paper, we address this challenge by first consolidating different procedures of explainability into a single definition based on the joint probability distribution of categories and rationales. This unified definition is essential because the model should be capable of recognizing both true categories and true rationales rather than focusing solely on one aspect. However, the crux of explainability lies in utilizing rationales for category prediction. We propose a prompt-based model that predicts rationales in a photo in the first step and then utilizes these rationales to predict the category in the second step. Experimental results demonstrate that despite its interpretable nature, this approach achieves state-of-the-art performance in explainable classification, both on individual datasets and in zero-shot scenarios across multiple datasets.

The roadmap of the paper is as follows: In Section \ref{sec:2}, we conduct a literature review on Vision Language Models, fine-tuning techniques, Explainability, and Visual Reasoning. Section \ref{sec:3} provides a detailed explanation of our proposed method. In Section \ref{sec:4}, we analyze our experiments and present our results. Finally, Section \ref{sec:5} concludes the paper by summarizing our contributions. We plan to share our code after the paper's publication.

\section{Related Work}\label{sec:2}

\subsubsection{Vision Language Models}
Vision-language models (VLMs) have rapidly emerged, creating a strong connection between visual information and natural language \cite{du2022survey}. These models learn to represent both images and their corresponding textual descriptions, unlocking a diverse set of capabilities, from generating detailed image captions \cite{wang2022git, 9087226} that capture scene dynamics and emotions to answering complex questions about visual content \cite{wang2022git, wang2022image, de2023visual}. Their capability to bridge the gap between vision and language opens doors for significant advancements in various fields. VLMs offer unmatched potential for applications like image search \cite{lahajal2024enhancing, Liu_2021_ICCV, moro2023efficient}, human-computer interaction \cite{zhou2023language}, and even creative content generation \cite{koh2024generating, ko2023large}. 

\subsubsection{Explainability}
While Deep Neural Networks (DNNs) excel at image classification, their "black box" nature poses challenges in applications requiring understanding the "why" behind predictions. This lack of explainability reduces trust and interpretability \cite{xu2019explainable}, particularly in sensitive domains like healthcare \cite{tjoa2020survey, stiglic2020interpretability, dave2020explainable}, autonomous vehicles \cite{atakishiyev2021explainable, glomsrud2019trustworthy, zhou2023vision}, and legal decision-making \cite{richmond2024explainable}. While traditional explainability methods like saliency maps \cite{simonyan2013deep} and gradient-based \cite{ribeiro2016should, selvaraju2017grad, Chen_2022_ACCV} methods offer valuable insights into local feature importance within DNNs, they often struggle to capture the broader reasoning and decision-making processes of complex models \cite{boccignone2019problems}. Recent advancements in explainable AI have explored more holistic approaches, leveraging the power of large language models (LLMs) to generate textual explanations \cite{menon2022visual, Naeem_2023_CVPR}. These methods make the internal mechanism of DNNs more human-understandable by translating model decisions into natural language narratives.

\subsubsection{Visual Attributes \& Visual Reasoning}
Visual attributes of objects have caught attention in recent research in computer vision \cite{patterson2016coco, krishna2017visual, pham2021learning, escorcia2015relationship}. However, the previous models were not guaranteed to focus on the main visual attributes of the object itself to make predictions. This drawback improves the possibility of wrong classification caused by attention to the background of the object \cite{singla2021salient, mao2023doubly}. Our work is trained on the visual attributes (rationales) of objects as well as the class of objects, which has led to superior performance compared to the other baselines.

\subsubsection{Prompt-Tuning}
Fully fine-tuning large-scale deep learning models like Transformer \cite{ding2023parameter, xin2024parameter, hu2021lora} needs too much resources and time and is not feasible in some cases. So, one method that is used in recent research papers is prompt-tuning \cite{lester2021power, liu2021p}, where a specific  number of prompts are added to some layer of the transformer. Specifically, the visual prompts are used in various computer vision tasks to fine-tune the visual network, such as VIT \cite{dosovitskiy2021an}, by introducing a smaller number of parameters compared to the model itself \cite{jia2022visual}. In our work, we have used the prompting technique to extract visual rationales from images.

\section{Method}\label{sec:3}
In this section, we first provide problem formulation (subsection \ref{subsec:3.1}), then we briefly overview the architecture of CLIP (subsection \ref{subsec:3.2}), and finally, we give a detailed explanation of our model, including training and inference (subsection \ref{subsec:3.3}).

\subsection{Problem Formulation}\label{subsec:3.1}
We introduce a general definition of explainability in an object recognition task, regardless of the base model we use.
\begin{definition}
    Consider an Image, denoted as $I$, belonging to category $c$, with rationales represented as $\cb{r_i}_{i=1}^{m}$ within the image corresponding to that category. An Explainable Model is expected to assign a high value to the joint probability of the true category and true rationales given the image, expressed as $P(c,\cb{r_i}_{i=1}^{m}\left|I\right.)$. 
\end{definition}
While the above definition aligns with human intuition, it warrants further clarification. For an explainable model, assigning a high value to $P(c,\cb{r_i}_{i=1}^{m}\left|I\right.)$ is necessary but no sufficient. It is necessary to ensure accurate prediction of both the true categories and rationales. However, it is not sufficient on its own because the method by which categories and rationales are predicted also matters. There are three potential scenarios to consider regarding this joint distribution:
\begin{itemize}
    \item If we assume independence between category and rationales given the image, i.e., $P(c,\cb{r_i}_{i=1}^{m}\left|I\right.) = P(c\left|I\right.)P(\cb{r_i}_{i=1}^{m}\left|I\right.)$, this approach is flawed as the essence of explainability lies in utilizing rationales to inform category prediction, indicating their dependence.

     \item Alternatively, if we first predict the category and then the rationales based on it, i.e., 
 $P(c,\cb{r_i}_{i=1}^{m}\left|I\right.) = P(c\left|I\right.)P(\cb{r_i}_{i=1}^{m}\left|c,I\right.)$, this approach is misguided as the purpose of rationales is to aid in category classification, not the other way around.
     
     \item The only reasonable approach is first to identify rationales in the image and then predict the category based on them. Hence, $P(c,\cb{r_i}_{i=1}^{m}\left|I\right.) = P(\cb{r_i}_{i=1}^{m}\left|I\right.)P(c\left|\cb{r_i}_{i=1}^{m},I\right.)$.
\end{itemize}

In the subsection \ref{subsec:3.3}, we elaborate on how these two probability distributions, $P(\cb{r_i}_{i=1}^{m}\left|I\right.)$ and $P(c\left|\cb{r_i}_{i=1}^{m},I\right.)$, can be modeled using CLIP.

\subsection{CLIP Overview}\label{subsec:3.2}
Consider a set of (image, text) pairs denoted by $\left\{I_i, t_i \right\}_{i=1}^{N}$. CLIP \cite{pmlr-v139-radford21a} employs a Text Encoder, $\mathcal{T}:\text{text}\rightarrow \mathbb{R}^{d}$ and an Image Encoder $\mathcal{I}:\text{image}\rightarrow \mathbb{R}^d$, which convert text and image into embedding vectors in multi-modal space, represented as, $\mathbb{R}^d$.
\subsubsection{Training}
CLIP utilizes the following contrastive loss in training:
\footnotesize
\begin{align}
\mathcal{L}_\text{CLIP}\left(\left\{I_i\right\}_{i=1}^N, \left\{t_i\right\}_{i=1}^{N}\right) = -\frac{1}{N}\sum_{i,j} y_{ij}\log \frac{\exp(\mathcal{T}(t_j)^T \mathcal{I}(I_i))}{\sum_{k}\exp\left(\mathcal{T}(t_k)^T \mathcal{I}(I_i)\right)}
\end{align}
\normalsize
where
\begin{align}
    y_{i,j} = \begin{cases}
        1 & \text{$I_i$ and $t_j$ match.}\\
        0 & \text{otherwise.}
    \end{cases}
\end{align}
\subsubsection{Inference} Consider we have images, $\left\{I_i\right\}_{i=1}^{M}$, and texts, $\left\{t_j\right\}_{j=1}^{N}$. During inference, we need to check the cosine similarity of text embeddings and image embeddings to find the most probable text for each image:
\begin{align}
    \hat{t}_{i} = \text{argmax}_{j}\quad \mathcal{T}(t_j)^T \mathcal{I}(I_i)
    \quad ,\forall i=1,2,\dots,M
\end{align}

\begin{figure*}[tb]
    \centering
    \begin{subfigure}{\textwidth}
        \includegraphics[width=\textwidth]{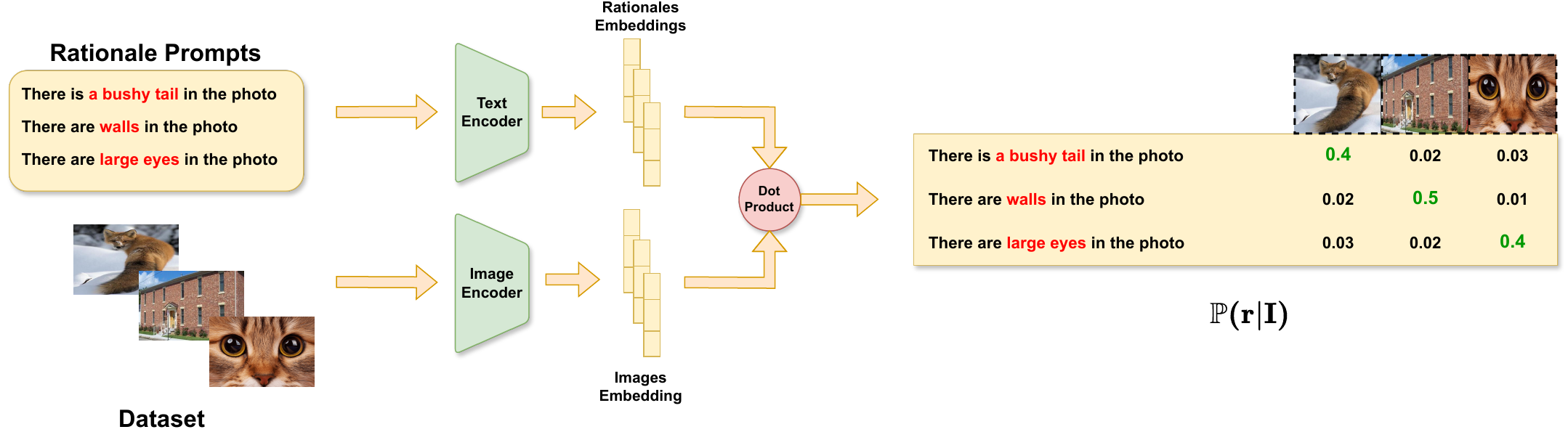}
        \caption{Stage 1}
    \end{subfigure}
    \vfill
    \begin{subfigure}{\textwidth}
        \includegraphics[width=\textwidth]{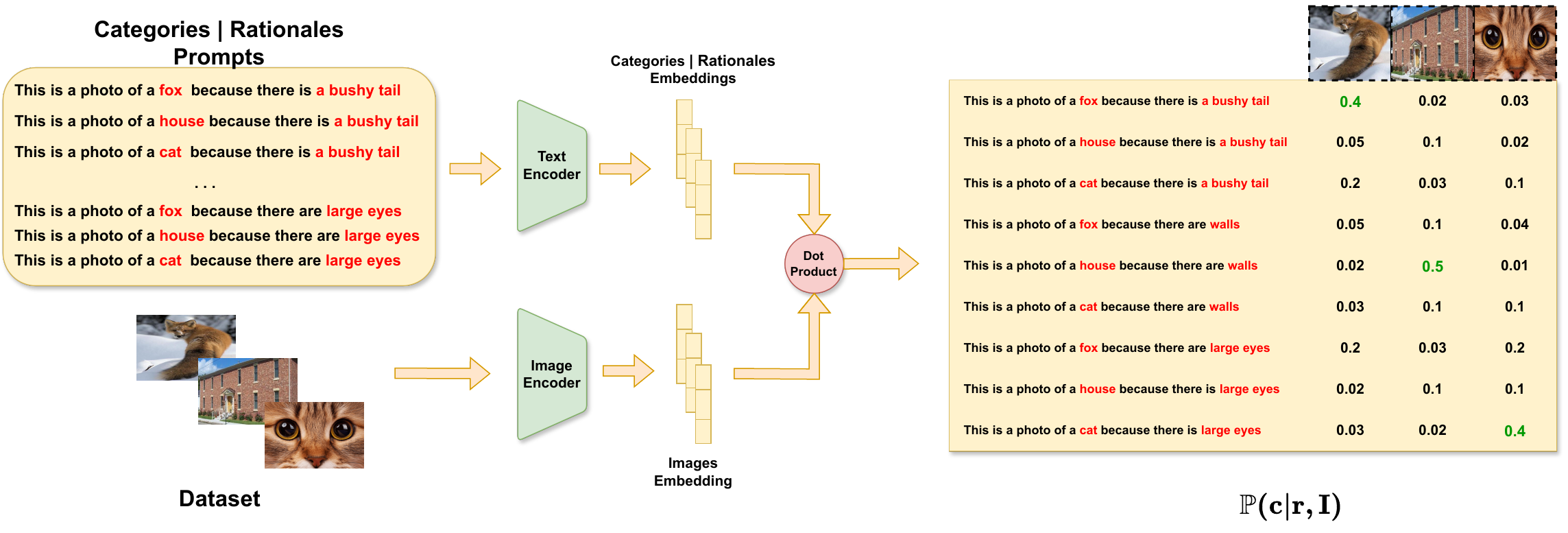}
        \caption{Stage 2}
    \end{subfigure}
    \caption{In the first step (a), the model utilizes Prompt$_{R}$to calculate the distribution of rationales. Then in Stage 2 (b), it identifies the distribution of categories conditioned on rationales using Prompt$_{c|R}$. Notably, the model's ability to detect categories is informed by the identified rationales, providing an explanation for its choices.}
    \label{fig:enter-label}
\end{figure*}

\subsection{Model}\label{subsec:3.3}
We first model the joint distribution of category and rationales as discussed in subsection \ref{subsec:3.1}, then we discuss different prompt-tuning methods that we considered for CLIP fine-tuning, followed by an explanation of our training and inference schemes.
\subsubsection{Joint probability distribution of category and rationales}
As discussed in subsection \ref{subsec:3.1}, our task entails modeling two probability distributions, $P(\cb{r_i}_{i=1}^{m}\left|I\right.)$ and $P(c\left|\cb{r_i}_{i=1}^{m},I\right.)$ using CLIP. Here, we introduce a method based on prompt engineering.

To model the distribution of rationales given an image, we devise the following text prompt:
\begin{itemize}
    \item Prompt$_{R}\doteq$ \footnotesize\emph{There are \{$r_1$\} and \{$r_2$\} and ... and \{$r_m$\} in the photo.}
\end{itemize}
where $R = \cb{r_i}_{i=1}^{m}$ represents rationales in image $I$. With the above text prompt and considering the CLIP Architecture (subsection \ref{subsec:3.2}), the probability of rationales $\cb{r_i}_{i=1}^{m}$ becomes:
\begin{align}
    P(\cb{r_i}_{i=1}^{m}\left|I\right.) = \text{\emph{Softmax}}\pran{\mathcal{T}(\text{Prompt}_{R})^T \mathcal{I}(I)}
\end{align}
where \emph{Softmax} is computed over all possible rational prompts in the dataset.

Now, focusing on $P(c\left|\cb{r_i}_{i=1}^{m},I\right.)$, we use the following text prompts to model this conditional distribution:
\begin{itemize}
    \item Prompt$_{c|R}\doteq$ \footnotesize\emph{This is a photo of a \{$c$\} because there is \{$r_1$\} and ... and \{$r_m$\}.}
\end{itemize}

We consider this representation of conditioning as an effective approach, and our ablation studies in subsection \ref{subsec:4.4} validate this choice. Hence, the conditional distribution of categories becomes:
 \begin{align}
    P(c\left|\cb{r_i}_{i=1}^{m},I\right.) = \text{\emph{Softmax}}\pran{\mathcal{T}(\text{Prompt}_{c|R})^T \mathcal{I}(I)}
\end{align}
where \emph{Softmax} are taken among all possible conditional prompts in the dataset, including all permutations of categories and rationales. This comprehensive approach aims to train our model in an autoregressive manner, which we'll discuss in training details soon.

\subsubsection{Prompt-Tuning}
To fine-tune CLIP, we explore two prompt-tuning methods introduced by \cite{mao2023doubly, jia2022visual}.
Shallow Prompt \cite{mao2023doubly} involves simply appending $K$ learnable prompts to the input of the vision transformer (Image Encoder):

\begin{align}
    e(I) \leftarrow \pran{e(I),p_0,p_1,\dots,p_K}.
\end{align}

Here, $e(I)\in \mathbb{R}^{L\times d}$ represents the input image to the vision transformer, where $L$ denotes the number of image tokens and $d$ represents the transformer embedding dimension. Additionally, $\cb{p_k}_{k=1}^{K} \in \mathbb{R}^{d}$ are learnable prompts appended to the image tokens. We employ this type of prompt-tuning for small datasets such as CIFAR-100.

Deep Prompts\cite{jia2022visual} append learnable prompts to intermediate vision transformer layers, too:
\begin{align}
    e^{l}(I) &\leftarrow \pran{e^l(I),p_0^l,p_1^l,\dots,p_K^l}\\
    e^{l+1}(I) &\leftarrow \text{VT}^{l}\pran{e^{l}(I)}
\end{align}
Here, $x^l$ and $\cb{p_k^{l}}_{k=1}^{K}$ represent the input and learnable prompts at layer $l$, respectively. $\text{VT}^{l}$ denotes the vision transformer block at layer $l$. This approach to fine-tuning is suitable for large datasets such as ImageNet.

These prompt-tuning methods enhance CLIP's ability to adapt to specific datasets and improve performance.

\subsubsection{Training} Let $\mathcal{D} = \left\{I_i, \left\{r_{j}^{(i)}\right\}_{j=1}^{m_i}, c_i \right\}_{i=1}^{N}$ denote our dataset, where $R^{(i)}=\cb{r_{j}^{(i)}}_{j=1}^{m_i}$ and $c_i$ represent the rationales and category of image $I_i$ respectively.

To model step-by-step thinking, i.e., first predicting rationales and then the category, we train CLIP in an autoregressive manner. In the first step, we predict rationales for a given image. In the second step, we predict the category conditioned on the predicted rationales. Consequently, the training loss is composed as follows:
\begin{align}
\mathcal{L}_{\text{train}}
&= \mathcal{L}_{\text{CLIP}}\pran{\cb{I_{i}}_{i=1}^{N}, \cb{\text{Prompt}_{R^{(i)}}}_{i=1}^{N}}\\
&+ \mathcal{L}_{\text{CLIP}}\pran{\cb{I_{i}}_{i=1}^{N}, \cb{\text{Prompt}_{c^{(i)}|R^{(j)}}}_{i,j=1}^{N}}
\end{align}

Where $\mathcal{L}_{\text{CLIP}}$ was defined in subsection \ref{subsec:3.2}. The first term represents the cross-entropy loss for rationale probability distribution, i.e., $\mathbb{E}_{\mathcal{D}}\left[-\log P(r|I)\right]$ and the second term is the cross-entropy loss for the conditional distribution of categories given rationales, i.e., $\mathbb{E}_{\mathcal{D}}\left[-\log(P(c|r,I)\right]$.

\subsubsection{Evaluation} For a given image $I$ whose rationales and category we aim to predict, we determine the rationales and category that maximize the joint distribution, $P(c,r|I)$, as follows:
\begin{align}
    c, r = \text{argmax}_{c,r}\, P(r|I)P(c|r,I)
\end{align}
In every Image can be multiple rationales, so we select the top $k_I$ of pair $c,\, r$, which maximizes joint distribution, where $k_I$ is a hyperparameter and depends on image $I$ \footnote{In inference, we don't access to ground truths, in this situation we set $k_I$ as the average number of rationales per image in the dataset, i.e., $K_I = \frac{1}{N}\sum_{i=1}^{N} m_i$, where $\cb{r_{j}^{i}}_{j=1}^{m_i}$ is the rationales for image $I_i$ in the dataset $D$.}, because the number of rationales in ground truth for each image, could vary. Finally, we use max voting among top $K_I$ sections for category prediction.

\section{Experiments}\label{sec:4}
We conducted extensive experiments on six diverse datasets to assess the effectiveness of our approach. This involved outperforming previous methods in explainable image classification across single datasets and extending to zero-shot settings. Additionally, we performed an ablation study to analyze the contributions of different model components.

\subsection{Experiments Setup}\label{subsec:4.1}

\subsubsection{Datasets}
\begin{figure}[t]
    \centering
    \includegraphics[width=\linewidth]{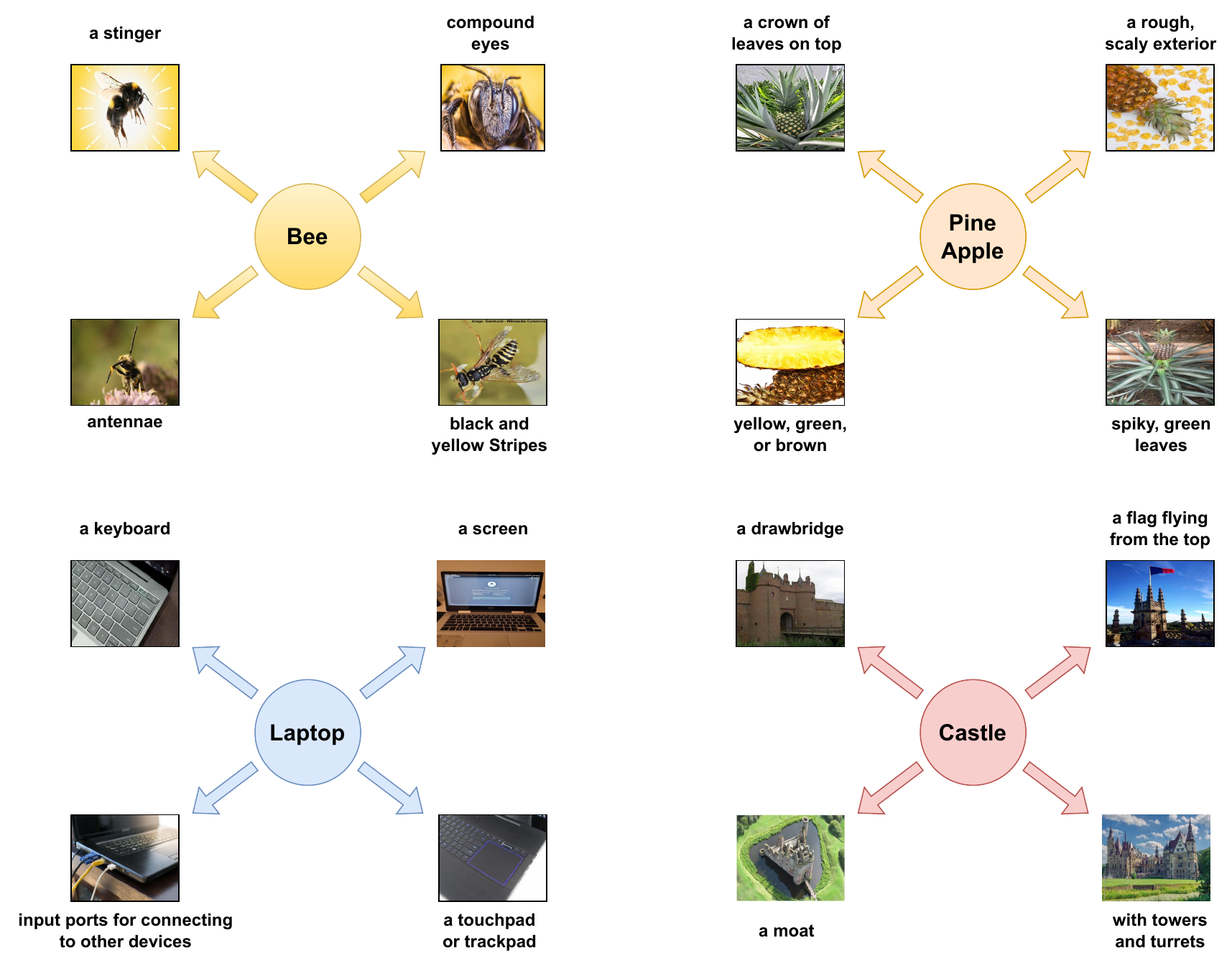}
      \caption{Examples of the generated rationales and their corresponding images in the DROR ImageNet dataset \cite{mao2023doubly}. The central words represent categories, while the surrounding words depict the corresponding rationales. Additionally, we provide one example image, retrieved through Google search, for each category and rationale.}
    \label{fig:2}
\end{figure}

We utilize DROR datasets prepared by \cite{mao2023doubly}, which are publicly available. Their dataset generation process is outlined as follows:
\begin{enumerate}
    \item Choose categories names of a dataset, e.g., CIFAR100 \cite{krizhevsky2009learning}.
    \item Ask GPT3 \cite{NEURIPS2020_1457c0d6} with the prompt, \emph{What are useful visual features for distinguishing a \{category name\} in a photo?}
    \item Collect attributes predicted by GPT and then search for images via Google Image API using the query, \emph{\{category name\} which has \{attribute name\}}
\end{enumerate}
They repeat above process for datasets CIFAR-10 \cite{krizhevsky2009learning}, CIFAR-100 \cite{krizhevsky2009learning}, Caltech-101 \cite{li_andreeto_ranzato_perona_2022}, Food-101 \cite{bossard14}, SUN \cite{5539970} and ImageNet \cite{ILSVRC15}. More details can be found at \cite{mao2023doubly}.
Figure \ref{fig:2} shows some example data obtained by this procedure. It should be noted that each image has one category and one rationale, so the number of rationales for each image in our formulization in section \ref{sec:3} is one.

\subsubsection{Hyperparameter Setup}
Following the approach of \cite{mao2023doubly}, we adopt shallow prompt-tuning for small datasets (as discussed in subsection \ref{subsec:3.3}) to efficiently fine-tune CLIP, prevent overfitting, and maintain zero-shot performance. For large datasets such as ImageNet, we employ deep prompt-tuning. Across all datasets, we utilize the CLIP-L/14 model, except for ImageNet, where we use the CLIP-B/32 model due to its larger size. Additionally, we conduct training on a single Nvidia Tesla V100 GPU. Table \ref{tab:1} illustrates our training setup.
Finally, in the evaluation, we consider the top 5 selections as explained in subsection \ref{subsec:3.3}.

\subsubsection{Metrics}
To evaluate results, we need to consider metrics that tell us how much our model is good in explainable classification. These metrics are introduced by \cite{mao2023doubly} along with datasets. There are 4 metrics:
\begin{itemize}
    \item RR: right category and right rationale\\
    \item RW: right category and wrong rationale\\
    \item WR: wrong category and right rationale\\
    \item WW: wrong category and wrong rationale\\    
\end{itemize}
In ideal circumstances, the RR metric would be as high as possible, indicating accurate predictions in both category and rationale. Conversely, we aim for the other metrics (RW, WR, WW) to be as low as possible, signifying minimal errors in the model's predictions. The sum of these 4 metrics must be 100\%.

\begin{table}[tb]
  \footnotesize
  \caption{Training setup.}
  \label{tab:1}
  \begin{tabular}{llll}
    \toprule
    Dataset & CLIP Model & Prompt-Tuning & Learnble Prompts\\
    \midrule
    CIFAR-10 & CLIP-L/14 & Shallow & 3\\
    CIFAR-100 & CLIP-L/14 & Shallow & 3\\
    Food-101 & CLIP-L/14 & Shallow & 3\\
    Caltech-101 & CLIP-L/14 & Shallow & 3\\
    SUN & CLIP-L/14 & Shallow & 100\\
    ImageNet & CLIP-B/32 & Deep & 30\\
  \bottomrule
\end{tabular}
\end{table}

\subsection{Baselines}
To benchmark our results and demonstrate the effectiveness of our model, we consider two other baselines: \textbf{CLIP} \cite{pmlr-v139-radford21a} and \textbf{DROR} \cite{mao2023doubly}.

\textbf{CLIP (Contrastive Language-Image Pre-training)}. In vanilla CLIP, the input text prompt is \emph{This is a photo of a \{category\}}. So, it doesn't involve rationales in fine-tuning. For a fair comparison, we use the same experiment setups as discussed in subsection \ref{subsec:4.1}.

\textbf{DDOR (Doubly Right Object Recognition).} This recent work has improved vanilla CLIP in explainable object recognition with prompt engineering. Their input text prompt is \emph{This is a photo of \{category\} because there is \{rationale\}.} Again, we use the same experiment setups as explained in subsection \ref{subsec:4.1} for a fair comparison.

\subsection{Results}
Extensive experiments on six diverse datasets demonstrate the strengths of our approach.
\subsubsection{Single Dataset Performance}
 Results on individual datasets are reported in Tabel \ref{tab:2}. Our model achieves state-of-art performance across five datasets. On CIFAR-10, due to the simplicity of classes, Autoregressive modeling does not show its effectiveness. This is an important observation that as datasets become larger and richer, our approach, which is based on step-by-step thinking, becomes more effective. For instance, on ImageNet, we observe a 144\% improvement over DROR.

 \subsubsection{Zero-Shot Performance} In the zero-shot experiment, we consider a training dataset for our model and a separate testing dataset for zero-shot evaluation. To ensure the reliability of results, we select the training dataset to be more general than the testing dataset. Results are presented in Table \ref{tab:3}. Our model achieves state-of-the-art performance in zero-shot settings. Across all setups, our model outperforms the previous two baselines, except for CIFAR-10, where the performance is close. The superiority of our model becomes more evident when the testing dataset is more domain-specific, such as Food-101, where it exclusively includes food-related images. Regardless of the training dataset, our method demonstrates significantly better explainability compared to DROR and CLIP on the Food-101 dataset. This capability can enhance the trustworthiness and generalizability of Vision Language Models in specialized domains.

 \subsubsection{Saliency maps} 
\begin{figure*}[t]
    \centering
    \includegraphics[width=\linewidth, height=8cm]{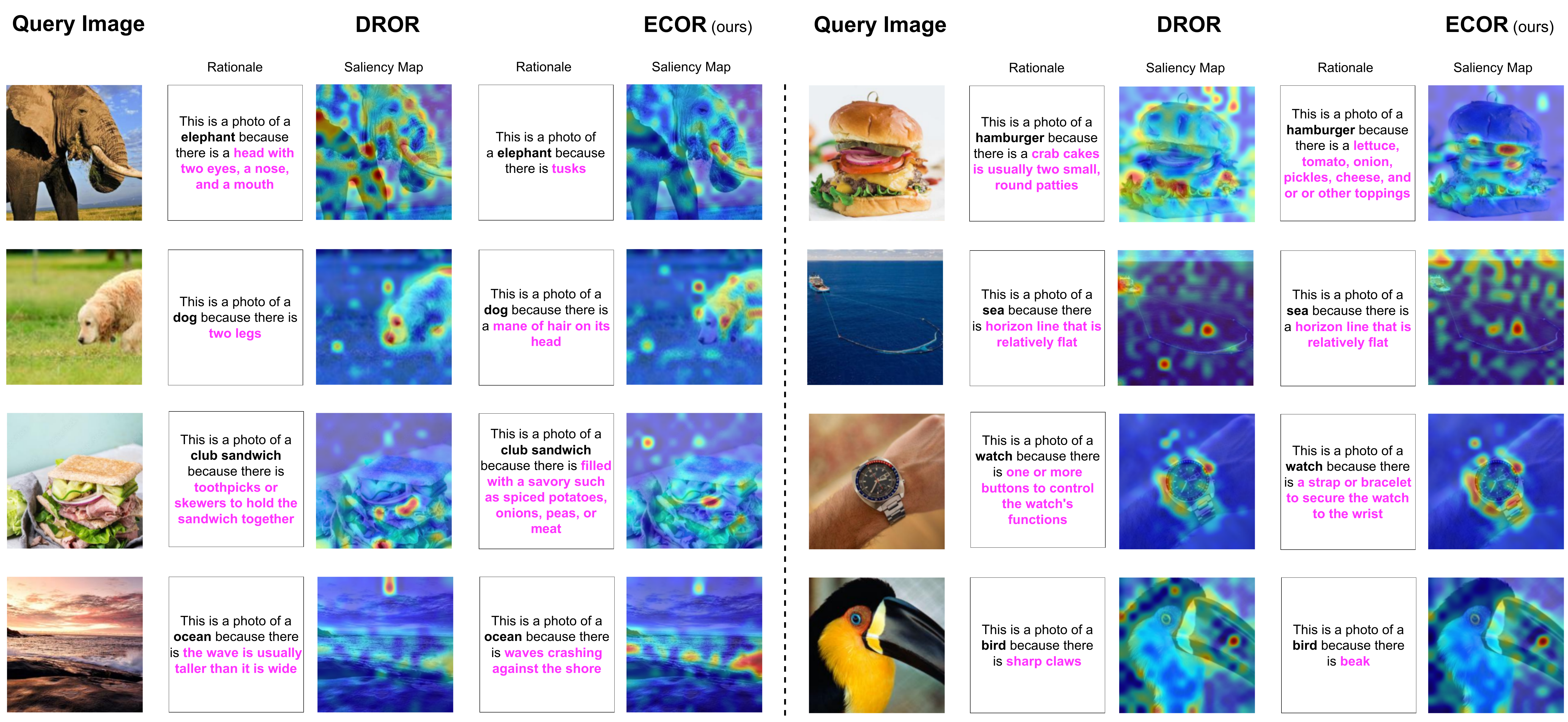}
      \caption{Visualization of saliency maps. Images are sampled from DROR datasets \cite{mao2023doubly}. In columns 2, 4, 7, and 9, we show the rationales produced by the model to explain the prediction. In columns 3, 5, 8, and 10, we show the saliency map \cite{Chen_2022_ACCV} that the models used to produce the prediction and rationales. Our method produces the correct category and rationales by attending to the appropriate parts of the images. Moreover, our model is more robust against being deceived by objects that are similar to the correct rationales but incorrect. This is clear in the ocean example (row 4, column 1), as our model correctly detected the waves, in contrast to the baseline model that focused on the clouds.}
    \label{fig:3}
\end{figure*}

The provided saliency maps, shown in Figure \ref{fig:3}, offer insight into the model's decision-making process by highlighting the significant image regions influencing the classification. Through this visualization, it becomes apparent that our model generates more accurate rationales compared to the baseline model. Moreover, it demonstrates a keen ability to focus on the relevant parts of the images, contributing to its superior performance in explainable object recognition.


\begin{table}[tb]
  \footnotesize
  \caption{Comparison of explainable object recognition performance across six datasets. Setups for these experiments are discussed in the subsection \ref{subsec:4.1}. Our model outperforms other baselines with a considerable gap. As the dataset becomes larger, the effect of autoregressive modeling, which we used, becomes clearer.}
  
  \label{tab:2}
  \begin{tabular}{llalll}
    \toprule
    Dataset & Model & RR $\uparrow$ & RW $\downarrow$ & WR $\downarrow$ & WW $\downarrow$\\
    \midrule
    \multirow{3}{*}{CIFAR-10} & CLIP & 42.57 & 44.52 & 7.06 & 5.84\\
    & DROR & \textbf{70.82} & 18.25 & 6.32 & 4.62\\
    & ECOR (Ours) & 65.69 & 24.09 & 4.87 & 5.35\\
    
    \midrule
     \multirow{3}{*}{CIFAR-100} & CLIP & 6.43 & 63.71 & 7.73 & 22.13\\
     & DROR & 22.27 & 44.61 & 9.97 & 23.14\\
     & ECOR (Ours) & \textbf{29.49} & 44.26 & 7.56 & 18.7\\

    \midrule
     \multirow{3}{*}{Food-101} & CLIP & 5.73 & 70.07 & 4.30 & 19.91\\
     & DROR & 25.25 & 51.83 & 5.83 & 17.08\\
     & ECOR (Ours) & \textbf{30.25} & 46.11 & 7.02 & 16.63\\

    \midrule
     \multirow{3}{*}{Caltech-101} & CLIP & 5.99 & 66.55 & 5.96 & 21.50\\
     & DROR & 23.64 & 52.43 & 5.86 & 18.06\\
    & ECOR (Ours) & \textbf{27.24} & 47.16 & 6.59 & 19.01\\

    \midrule
     \multirow{3}{*}{SUN} & CLIP & 0.94 & 24.27 & 11.21 & 63.58\\
     & DROR & 6.70 & 8.29 & 23.76 & 61.24\\
     & ECOR (Ours) &  \textbf{17.01} & 23.25 & 14.93 & 44.81\\

    \midrule
     \multirow{3}{*}{ImageNet} & CLIP & 0.68 & 42.69 & 3.87 & 52.76\\
     & DROR & 3.63 & 21.70 & 7.66 & 25.34\\
     & ECOR (Ours) & \textbf{8.87} & 30.35 & 7.04 & 53.74\\
    \bottomrule
  \end{tabular}
\end{table}

\begin{table*}[tb]
  \footnotesize
  \centering
  \caption{Comparison of explainable object recognition zero-shot performance across six datasets. As the testing dataset narrows down to more specific domains like Food-101, the superiority of our models becomes increasingly evident in zero-shot transferability.}
  \label{tab:3}
  \begin{tabular}{lllalll}
    \toprule
    Training Dataset & Testing Dataset & Model & RR $\uparrow$ & RW $\downarrow$& WR $\downarrow$ &WW $\downarrow$\\
    \midrule
    \multirow{3}{*}{CIFAR-100} & \multirow{3}{*}{CIFAR-10} & CLIP & 42.47 & 44.52 & 7.06 & 5.84\\
    & & DROR & \textbf{54.99} & 33.58 & 4.87 & 6.57\\
    & & ECOR (Ours) & 52.07 & 33.33 & 8.76 & 5.84\\

    \midrule
     \multirow{3}{*}{CIFAR-100} & \multirow{3}{*}{Food-101} & CLIP & 5.73 & 70.07 & 4.30 & 19.91\\
    & & DROR & 8.35 & 44.61 & 9.97 & 23.14\\
    & & ECOR (Ours) & \textbf{18.97} & 48.55 & 7.72 & 24.76\\

     \midrule
     \multirow{3}{*}{CIFAR-100} & \multirow{3}{*}{Caltech-101} & CLIP & 5.99 & 66.55 & 5.96 & 21.50\\
    & & DROR & 15.07 & 56.34 & 6.21 & 22.38\\
    & & ECOR (Ours) & \textbf{21.12} & 49.34 & 7.47 & 22.07\\
   \midrule
   \midrule
     \multirow{3}{*}{Caltech101} & \multirow{3}{*}{CIFAR-10} & CLIP & 42.47 & 44.52 & 7.06 & 5.84\\
    & & DROR & 49.63 & 39.17 & 5.60 & 5.60\\
    & & ECOR (Ours) & \textbf{53.53} & 34.31 & 5.11 & 7.06\\

   \midrule
     \multirow{3}{*}{Caltech101} & \multirow{3}{*}{CIFAR-100} & CLIP & 6.43 & 63.71 & 7.73 & 22.13\\
    & & DROR & 13.20 & 49.11 & 10.38 & 27.30\\
    & & ECOR (Ours) & \textbf{20.24} & 47.11 & 9.19 & 23.47\\

    \midrule
     \multirow{3}{*}{Caltech101} & \multirow{3}{*}{Food-101} & CLIP & 5.73 & 70.07 & 4.30 & 19.91\\
    & & DROR & 7.36 & 61.05 & 4.49 & 26.69\\
    & & ECOR (Ours) & \textbf{19.39} & 51.00 & 6.95 & 22.67\\
   \midrule
   \midrule
     \multirow{3}{*}{SUN} & \multirow{3}{*}{CIFAR-10} & CLIP & 42.47 & 44.52 & 7.06 & 5.84\\
    & & DROR & \textbf{49.00}& 40.04 & 5.78 & 5.18\\
    & & ECOR (Ours) & 46.47 & 42.09 & 5.84 & 5.60\\

    \midrule
     \multirow{3}{*}{SUN} & \multirow{3}{*}{CIFAR-100} & CLIP & 6.43 & 63.71 & 7.73 & 22.13\\
    & & DROR & 13.11& 49.63 & 8.32 & 28.93\\
    & & ECOR (Ours)& \textbf{18.03} & 47.92 & 8.26 & 25.79\\

    \midrule
     \multirow{3}{*}{SUN} & \multirow{3}{*}{Food-101} & CLIP & 5.73 & 70.07 & 4.30 & 19.91\\
    & & DROR & 8.94& 51.90 & 6.67 & 32.48\\
    & & ECOR (Ours) & \textbf{14.57} & 43.07 & 9.22 & 33.15\\

    \midrule
     \multirow{3}{*}{SUN} & \multirow{3}{*}{Caltech-101} & CLIP & 5.99 & 66.55 & 5.96 & 21.50\\
    & & DROR & 13.58 & 55.14 & 5.39 & 25.88\\
    & & ECOR (Ours) & \textbf{18.28} & 49.43 & 7.19 & 25.09\\
   \midrule
   \midrule
     \multirow{3}{*}{ImageNet} & \multirow{3}{*}{CIFAR-10} & CLIP & 36.98 & 46.71 & 9.00 & 7.30\\
    & & DROR & \textbf{38.68} & 43.80 & 9.25 & 8.27\\
    & & ECOR (Ours) & 36.74 & 41.85 & 8.52 & 12.90\\

    \midrule
     \multirow{3}{*}{ImageNet} & \multirow{3}{*}{CIFAR-100} & CLIP & 5.23 & 59.93 & 7.30 & 27.53\\
    & & DROR & 15.67 & 39.89 & 9.51 & 55.57\\
    & & ECOR (Ours)& \textbf{16.25} & 37.31 & 9.33 & 37.10\\

    \midrule
     \multirow{3}{*}{ImageNet} & \multirow{3}{*}{Food-101} & CLIP & 5.48 & 63.57 & 5.41 & 25.53\\
    & & DROR & 8.31 & 46.59 & 5.97 & 39.12\\
    & & ECOR (Ours) & \textbf{10.90} & 37.30 & 8.35 & 43.45\\

    \midrule
     \multirow{3}{*}{ImageNet} & \multirow{3}{*}{Caltech-101} & CLIP & 4.51 & 65.19 & 5.42 & 24.15\\
    & & DROR & 16.71 & 45.42 & 7.66 & 30.20\\
    & & ECOR (Ours) & \textbf{17.09} & 39.19 & 8.42 & 35.31\\

    \midrule
     \multirow{3}{*}{ImageNet} & \multirow{3}{*}{SUN} & CLIP & 0.86 & 23.32 & 10.72 & 65.16\\
    & & DROR & 1.98 & 7.02 & 14.96 & 76.90\\
    & & ECOR (Ours) & \textbf{3.61} & 9.96 & 12.65 & 73.70\\
    \bottomrule
  \end{tabular}
\end{table*}

\begin{table*}[tb]
\centering
  \footnotesize
  \caption{Comparsion of Ablation Experiment. In AB1, training was just done on rationales. AB2 and AB3 consider approaches of vanilla CLIP \cite{pmlr-v139-radford21a} and DROR \cite{mao2023doubly}. AB4 keeps rationale prompts but inverse conditional prompts, i.e., \emph{<rational> because of <category>}. AB5 assumes independence between rationales and categories. AB6 first predicts the category and then predicts rationales based on the predicted category.
  Results show us for small datasets, CIFAR-10, CIFAR-100, Food-101, and Caltech-101, AB4 (Independece) performs better, although it is not explainable because rationales are not used for category prediction. For large datasets, SUN and ImageNet, ECOR outperforms other approaches by a large margin.}
  \label{tab:4}
  \begin{tabular}{llalll}
    \toprule
    Dataset & Approach & RR $\uparrow$ & RW $\downarrow$& WR $\downarrow$& WW $\downarrow$\\
    \midrule
    \multirow{6}{*}{CIFAR-10} & AB1 (Just Rationales) & 55.47 & 28.95 & 8.27 & 7.30\\ 
    & AB2 (Just Categories / CLIP) & 42.57 & 44.52 & 7.06 & 5.84\\
    & AB3 (Just Conditioning / DROR) & \textbf{70.82} & 18.25 & 6.32 & 4.62\\
    & AB4 (False Conditioning) & 64.23 & 24.57 & 6.81 & 4.38\\
    & AB5 (Independece) & 68.86 & 21.65 & 5.35 & 4.14\\
    & AB6 (Inverse ECOR) & 67.40 & 21.41 & 6.08 & 5.11\\
    & ECOR & 65.69 & 24.09 & 4.87 & 5.35\\
    
    \midrule
     \multirow{6}{*}{CIFAR-100} & AB1 (Just Ratinoales) & 16.52 & 37.6 & 16.28 & 29.60\\ 
     & AB2 (Just Categories / CLIP) & 6.43 & 63.71 & 7.73 & 22.13\\
     & AB3 (Just Rationales / DROR) & 22.27 & 44.61 & 9.97 & 23.14\\
     & AB4 (False Conditioning) & 26.34 & 42.26 & 7.71 & 19.69\\
     &  AB5 (Independece) & \textbf{35.39} & 40.51 & 7.59 & 16.52\\
     & AB6 (Inverse ECOR) & 22.59 & 45.68 & 10.35 & 21.37\\
     & ECOR & 29.49 & 44.26 & 7.56 & 18.7\\

    \midrule
     \multirow{6}{*}{Food-101} & AB1 (Just Ratinoales) & 26.06 & 44.78 & 9.08 & 20.08\\ 
     & AB2 (Just Categories / CLIP) & 5.73 & 70.07 & 4.30 & 19.91\\
     & AB3 (Just Rationales / DROR) & 25.25 & 51.83 & 5.83 & 17.08\\
     &AB4 (False Conditioning) & 29.41 & 45.55 & 6.95 & 18.09\\ 
     & AB5 (Independece) & \textbf{36.81} & 41.32 & 7.68 & 14.18\\ 
     &AB6 (Inverse ECOR) & 27.80 & 49.56 & 5.97 & 16.66\\ 
     & ECOR & 30.25 & 46.11 & 7.02 & 16.63\\

    \midrule
     \multirow{6}{*}{Caltech-101} & AB1 (Just Ratinoales) & 18.98 & 42.18 & 11.82 & 27.02\\ 
     & AB2 (Just Categories / CLIP) & 5.99 & 66.55 & 5.96 & 21.50\\
     & AB3 (Just Rationales / DROR) & 23.64 & 52.43 & 5.86 & 18.06\\
     &AB4 (False Conditioning) & 26.86 & 45.62 & 7.06 & 20.46\\ 
     & AB5 (Independece) & \textbf{34.21} & 42.37 & 6.18 & 17.24\\ 
     &AB6 (Inverse ECOR) & 26.06 & 44.78 & 9.08 & 20.08\\ 
    & ECOR & 23.61 & 52.59 & 5.86 & 17.94\\

    \midrule
     \multirow{6}{*}{SUN} & AB1 (Just Ratinoales) & 2.79 & 5.61 &25.56 & 66.04\\ 
     & AB2 (Just Categories / CLIP) & 0.94 & 24.27 & 11.21 & 63.58\\
     & AB3 (Just Rationales / DROR) & 6.70 & 8.29 & 23.76 & 61.24\\
     & AB4 (False Conditioning) & 14.48 & 21.61 & 15.94 & 47.97\\ 
     &  AB5 (Independece) & 14.52 & 25.91 & 13.48 & 46.09\\ 
     & AB6 (Inverse ECOR) & 7.17 & 7.98 & 25.74 & 59.10\\ 
     & ECOR &  \textbf{17.01} & 23.25 & 14.93 & 44.81\\

    \midrule
     \multirow{6}{*}{ImageNet}  & AB1 (Just Ratinoales) & 0.76 & 45.93 & 1.86 & 51.45\\ 
     & AB2 (Just Categories / CLIP) & 0.68 & 42.69 & 3.87 & 52.76\\
     & AB3 (Just Rationales / DROR) & 3.63 & 21.70 & 7.66 & 25.34\\
     & AB4 (False Conditioning) & 2.65 & 20.99 & 8.44 & 67.92\\ 
     &  AB5 (Independece) & 1.17 & 23.98 & 10.50 & 64.35\\ 
     & AB6 (Inverse ECOR) & 4.42 & 26.65 & 9.77 & 59.16\\ 
     & ECOR & \textbf{8.87} & 30.35 & 7.04 & 53.74\\
    \bottomrule
  \end{tabular}
\end{table*}

\subsection{Ablation Study}\label{subsec:4.4}
In this subsection, we explore the impact of text prompts utilized for modeling the joint distribution of categories and rationales. Our ablation studies reveal that our model represents a generalization beyond previous baselines, including CLIP and DDOR. We evaluate six distinct prompt designs to determine their effectiveness.
\subsubsection{AB1 (Just Rationales)}
In this experiment, we just train on $\text{Prompt}_{R}$, i.e., \emph{"There is \{$r$\} in the photo."}. The results in Table \ref{tab:4} show improvement over vanilla CLIP, resulting in improved detection of rationales. However, it still struggles to connect the relation between rationales and categories, resulting in a significant gap compared to ECOR.

\subsubsection{AB2 (Just Categories / CLIP)}
    In this experiment, we just train on categories, i.e., prompt \emph{"This is a photo of a <c>"}. This is the same as vanilla CLIP.

\subsubsection{AB3 (Just Conditioning / DROR)}
In this experiment, we solely train on the condition prompt or $\text{Prompt}_{c|R}$, i.e., \emph{"This is a photo of a \{c\} because there is \{r\}"}. This setup mirrors the approach of DROR \cite{mao2023doubly}.

\subsubsection{AB4 (False Conditioning)}
 In this ablation, we keep $\text{Prompt}_{R}$ unchanged, but we set $\text{Prompt}_{c|R}$ as \emph{"There is \{r\} because this is a photo of a \{c\}"}. In some ways, we actually invert the conditioning, which contradicts our assumption that categories should be conditioned on rationales. As seen in Table \ref{tab:4}, compared to ECOR, the performance drops slightly on small datasets CIFAR-10, CIFAR-100, Food-101, and Caltech-101. However, the gap is considerable for large datasets such as SUN and ImageNet. This is because small datasets have simpler categories, making it easier for VLMs like CLIP to recognize them even with false conditioning. However, false conditioning leads to poor performance for large datasets with more complex categories.

Now, we consider the other two approaches for modeling joint distribution, independence of rationales \& categories and conditioning rationales on categories, which we denied in subsection \ref{subsec:3.1}.
    \subsubsection{AB5 (Independence)}
    Recall from subsection \ref{subsec:3.1} that this approach considers $P(c, \{r_i\}_{i=1}^{m} | I) = P(c | I) P(\{r_i\}_{i=1}^{m} | I)$. For the rationales distribution, we consider $\text{Prompt}_{R} = $ \emph{"There are \{$r_1$\} and \{$r_2$\} and $\dots$ and \{$r_m$\} in the photo."}, while for the category distribution, we consider $\text{Prompt}_c = $ \emph{"This is a photo of a \{$c$\}."}. This approach is dismissed because CLIP treats rationales as new classes rather than hints or explanations for category prediction. However, the results in Table \ref{tab:4} show that for small datasets like CIFAR-10, CIFAR-100, Food-101, and Caltech-101, the performance is improved compared to ECOR for two reasons. First, because of the separation of rationale and category prompts, CLIP is not forced to use rationales for category prediction. Second, for these small datasets, the number of categories and rationales is limited, resulting in good performance in simultaneously predicting both categories and rationales. However, the results dropped significantly for large datasets like SUN and ImageNet, especially for ImageNet. Here, the number of rationales is too high to independently predict both class and rationales. In contrast, ECOR simplifies the situation for CLIP by using rationales as hints for class prediction, thereby enhancing explainability, reliability, and performance.

\subsubsection{AB6 (Inverse ECOR)}
    In this final ablation, we investigate the scenario where we first predict categories and then predict rationales. Therefore, we consider $P(c, \{r_i\}_{i=1}^{m} | I) = P(c | I) P(\{r_i\}_{i=1}^{m} | c, I)$. We dismissed this approach because we use rationales as hints for a category, not inversely. Also, as shown in Table \ref{tab:4}, the performance dropped across all datasets compared to ECOR (except CIFAR-10, where performances are close). The gap increases when the dataset becomes larger and has more complicated categories.

\section{Conclusion}\label{sec:5}
In conclusion, our paper proposes a novel approach to enhance the explainability of large Vision Language Models (VLMs) such as CLIP. By introducing a unified mathematical definition of explainability based on the joint probability distribution of categories and rationales, we establish a foundation for our method. We develop a prompt-based model that predicts rationales in the first step and utilizes them for category prediction in the second step. Through extensive experiments on various datasets, including zero-shot scenarios, our approach achieves state-of-the-art performance in explainable classification. Despite the interpretability of our model, it maintains high accuracy, addressing the challenge of balancing prediction accuracy with the provision of meaningful rationales. Our work contributes to improving trust and accountability in critical domains by providing transparent and interpretable explanations for object recognition. Future research can explore extending our method to other categories of VLMs, such as generative models, and investigate its applicability in additional domains.

\bibliographystyle{IEEEtran}
\bibliography{refrences}

\end{document}